\documentclass[10pt]{IEEEtran}
\usepackage{amsmath,amsfonts}
\usepackage{array}
\usepackage{ragged2e} 
\usepackage[caption=false,font=normalsize]{subfig}
\usepackage{textcomp}
\usepackage{url}
\usepackage{verbatim}
\usepackage{graphicx}
\usepackage{cite}
\usepackage{tikz,xcolor} 
\usepackage{hyperref} 
\usepackage{booktabs}  
\usepackage[OT1]{fontenc}  
\usepackage{gensymb}      
\usepackage{mathrsfs}     
\usepackage{stfloats}     
\usepackage{amssymb}      
\usepackage{algorithm}   
\usepackage{algorithmic} 
\usepackage{layout}

\usepackage[explicit]{titlesec}
\titlespacing{\section}{0pt}{1.2ex plus .0ex minus .0ex}{.3ex plus .0ex}
\titlespacing{\subsection}{0pt}{1.2ex plus .0ex minus .0ex}{.3ex plus .0ex}
\usepackage[
    top=0.7in, 
    bottom=0.99in,
    left=0.608in,
    right=0.608in,
]{geometry}
\usepackage{colortbl}  
\definecolor{aviationColor}{rgb}{0.80, 0.616, 0.42}    
\definecolor{marineColor}{rgb}{0.65, 0.804, 0.79} 
\definecolor{mypurple}{rgb}{0.7176, 0.361, 0.906}

\begin{document}

\title{AI-Enhanced Distributed Channel Access for Collision Avoidance in Future Wi-Fi 8}

\vspace{-0.8cm}

\author{
  Jinzhe Pan\textsuperscript{\dag},
	Jingqing Wang\textsuperscript{\ddag}, 
  Yuehui Ouyang\textsuperscript{\dag},
  Wenchi Cheng\textsuperscript{\ddag}, and
  Wei Zhang\textsuperscript{\S}

  {\small 
  \textsuperscript{\dag}Honor Device Company Ltd, Shenzhen, China

  \textsuperscript{\ddag}State Key Laboratory of Integrated Services Networks, Xidian University, Xi’an, China

  \textsuperscript{\S}School of Electrical Engineering and Telecommunications, The University of New South Wales, Sydney, Australia
  
  panjinzhe@honor.com, jqwangxd@xidian.edu.cn, ouyangyuehui@honor.com, wccheng@xidian.edu.cn, w.zhang@unsw.edu.au
}

  \vspace{-25pt}


    
}

\maketitle

\pagestyle{empty} 
\thispagestyle{empty} 

\begin{abstract}
  The exponential growth of wireless devices and stringent reliability requirements of emerging applications demand fundamental improvements in distributed channel access mechanisms for unlicensed bands. Current Wi-Fi systems, which rely on binary exponential backoff (BEB), suffer from suboptimal collision resolution in dense deployments and persistent fairness challenges due to inherent randomness. This paper introduces a multi-agent reinforcement learning framework that integrates artificial intelligence (AI) optimization with legacy device coexistence. We first develop a dynamic backoff selection mechanism that adapts to real-time channel conditions through access deferral events while maintaining full compatibility with conventional CSMA/CA operations. Second, we introduce a fairness quantification metric aligned with enhanced distributed channel access (EDCA) principles to ensure equitable medium access opportunities. Finally, we propose a centralized training decentralized execution (CTDE) architecture incorporating neighborhood activity patterns as observational inputs, optimized via constrained multi-agent proximal policy optimization (MAPPO) to jointly minimize collisions and guarantee fairness. Experimental results demonstrate that our solution significantly reduces collision probability compared to conventional BEB while preserving backward compatibility with commercial Wi-Fi devices. The proposed fairness metric effectively eliminates starvation risks in heterogeneous scenarios. 
\end{abstract}


\section{Introduction}
The exponential deployment of smart devices and Internet-of-Things (IoT) systems has intensified spectrum contention in unlicensed bands. Meanwhile, emerging applications with low latency communication requirements impose stringent demands on wireless networks, driving the evolution toward Wi-Fi 8 with high reliability targets \cite{wifi8}.
Recognizing this paradigm shift, the IEEE 802.11 standards group has pivoted its focus from throughput enhancement to reliability assurance in next-generation wireless networks \cite{IEEE80211PAR}.
Unlike the licensed spectrum with centralized resource allocation, unlicensed bands necessitate decentralized coexistence among heterogeneous technologies (Wi-Fi, Bluetooth, LTE-LAA, etc.) under regulatory constraints. 
This unique environment demands distributed channel access (DCA) strategies that simultaneously achieve self-organization capability, fairness and collision mitigation.

Modern Wi-Fi systems implement DCA through carrier-sense multiple access with collision avoidance (CSMA/CA), which operates on two fundamental mechanisms: listen before talk (LBT) and binary exponential backoff (BEB).
When attempting channel access, devices select random backoff counters within a dynamic contention window (CW), and suspend countdown during detected channel activity through LBT monitoring.
A number of research works have focused on optimizing CSMA/CA parameters, such as designing better CW adjustment strategy \cite{sun2014backoff}. 
However, the inherent randomness of BEB causes inefficient spectrum utilization in dense deployments \cite{dai2012unified} and fairness issues \cite{cagalj2005selfish}, which cannot satisfy the throughput and reliability requirements of future Wi-Fi 8.

Recent advancements in artificial intelligence (AI) across various domains have inspired researchers to explore its potential for optimizing Wi-Fi performance \cite{wifiMeetsMl}.
One direct application is to implement deep reinforcement learning (DRL) to intelligently adjust CSMA/CA parameters, such as adaptive CW tuning \cite{wydmanski2021contention, kumar2021adaptive}. 
However, these methodologies remain constrained by fundamentally limitations of random access protocols, failing to fully exploit AI's decision-making capabilities.
Emerging frameworks attempt to transcend these limitations through architectural innovations.
Valcarce {\emph{et al.}} \cite{9432398} developed a joint optimization framework for media access control (MAC) signaling and channel access policies, though its fairness guarantees with legacy CSMA/CA devices require further validation.
The adoption of multi-agent reinforcement learning (MARL) addresses cooperative access challenges:
Guo {\emph{et al.}} \cite{guo2022multi} utilized a QMIX-based centralized training with distributed execution (CTDE) architecture for slot-level transmission control in homogeneous networks enforcing fairness through priority learning. The microsecond-level decision latency requirement poses significant challenges for power-constrained devices.
Subsequent work by Yu et al. \cite{yu2021multi} further investigated and extended to heterogeneous networks using $\alpha$-fairness criteria, which may not inherently align with CSMA/CA's native fairness mechanisms.

Beyond throughput optimization through collision mitigation, a critical requirement for modern DCA strategies lies in maintaining backward compatibility with legacy devices. 
To address these dual challenges, we propose a solution by combining
\emph{(i)} the multi-agent proximal policy optimization (MAPPO) framework \cite{NEURIPS2022_9c1535a0}, enabling cooperative learning among stations, with 
\emph{(ii)} the reward-constrained policy optimization (RCPO) method \cite{tessler2018reward}, incorporating fairness as a learnable constraint. 
The key contributions are as follows:

\textbullet\ We propose a novel backoff reselection mechanism triggered after access deferral events, ensuring compatibility with CSMA/CA while fully exploiting AI-driven optimization of channel access strategies.

\textbullet\ We introduce a new fairness evaluation metric specifically designed for compatibility with enhanced DCA mechanisms.

\textbullet\ We propose a fairness-constrained MAPPO algorithm incorporating temporal channel information (cumulative idle time since last neighbor acknowledgement) to mitigate collisions and improve throughput while gauranteeing fairness.

The rest of the paper is organized as follows. The system model and the fairness-constrained optimization problem are introduced in Section \ref{sec2}. The details of the proposed FC-MAPPO are provided in Section \ref{sec3}. Simulation results are shown in Section \ref{sec4}. The conclusions are made in Section \ref{sec5}.

\section{System Model\label{sec2}}
This section introduces an innovative backoff selection mechanism with quantifiable fairness metrics, formulating the DCA problem as a constrained MARL task.

\begin{figure*}[!ht]
  \centering
  \includegraphics[width = 1\textwidth]{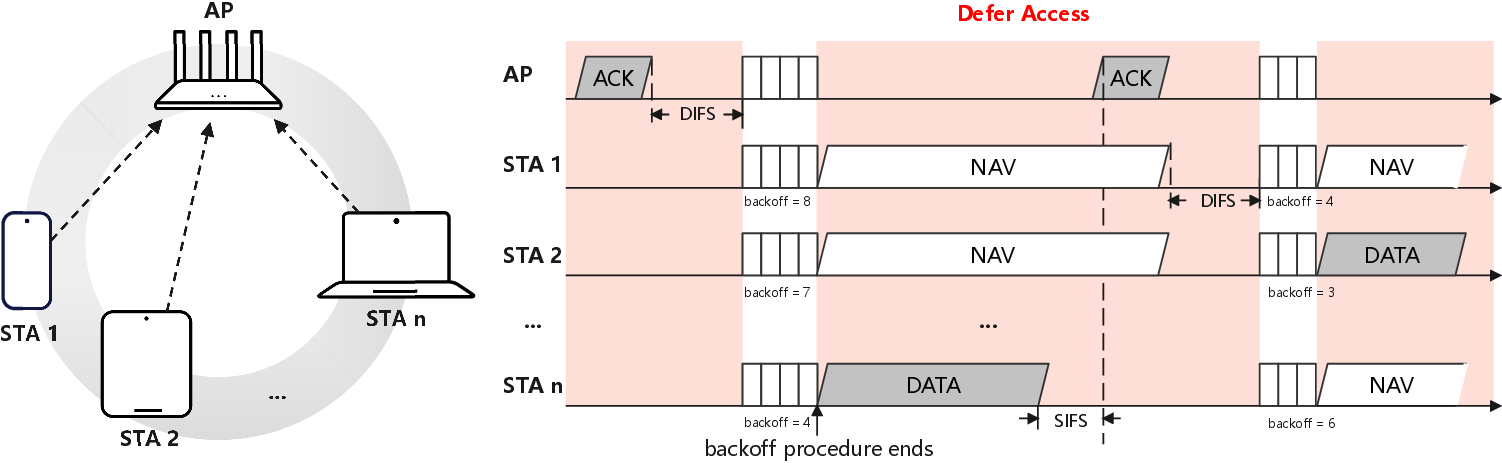}
  \vspace{-1.3em}
  \caption{Distributed access in a single BSS with multiple STAs sharing a common channel for uplink traffic to the AP.}
  \label{fig_1}
  \vspace{-1.3em}
\end{figure*}

\subsection{Distributed Channel Access}
We consider an IEEE 802.11ax basic service set (BSS) consisting of $n$ stations (STAs) transmitting uplink saturated traffic to their associated access point (AP), as illustrated in Fig. \ref{fig_1}.
All STAs employ finite queue buffers without implementing frame aggregation.
The request-to-send and clear-to-send (RTS/CTS) mechanism is disabled.
Traffic flows exclusively use access category best effort (AC\_BE) queues.
The network contains heterogeneous devices: legacy STAs and RL-enhanced STAs.
Both device types adhere to LBT principles: they must observe a distributed inter-frame space (DIFS) idle period before access attempts, and suspend backoff countdown during access deferral events (triggered by neighboring transmissions).
If a packet is correctly received, the STA will be replied with an acknowledgement (ACK) from the AP.
If the ACK is not received, the STA will retransmit the packet, doubling the CW size each time, until it either reaches $\text{CW}_{\max}$ or is reset to $\text{CW}_{\min}$ upon successful delivery.

\subsection{Defer Access-Based Backoff Selection Mechanism}
The core challenge of AI-driven channel access optimization lies in predicting collision probabilities across future CW slots and selecting the optimal slot with minimal collision risk.
However, in 802.11ax systems, each physical slot lasts only 9$\mu$s. Even aggregating multiple slots results in negligible durations, insufficient to observe meaningful network state variations for transmission/wait decisions.
By contrast, access deferral periods span hundreds of microseconds to milliseconds -- orders of magnitude longer than typical backoff counter durations. Subsequently, environmental dynamics (e.g., queue changes) predominantly occur during these access deferral intervals rather than backoff countdown phases.

We thus propose a defer access-based backoff selection mechanism that re-evaluates network states and dynamically adjusts backoff counters after each access deferral event.
Inspired by the Bianchi model \cite{bianchi}, which treats a backoff counter decrement interval as a "generic slot", we reinterpret the entire access deferral period as a single virtual slot. 
As illustrated in Fig. 2, our method operates between two extremes:
\emph{i)} Conventional CSMA/CA initiates channel access and select a backoff counter when a new packet is enqueued.
\emph{ii)} Slot-level decision mechanism like \cite{guo2022multi} makes transmission decision at each generic slot.
The proposed mechanism balances adaptability and practicality: it aligns decision-making with intrinsic MAC-layer deferral events, and avoids introducing impractical computational or timing overhead.

\begin{figure}[!t]
  \centering
  \includegraphics[width = 0.48\textwidth]{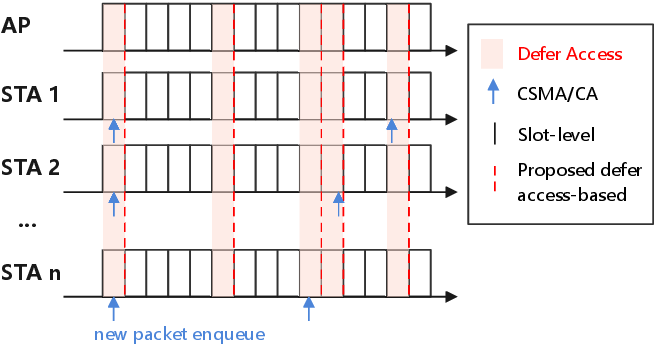}
  \vspace{-1.0em}
  \caption{Backoff selection timing comparison: Defer access-based mechanism vs. legacy CSMA/CA and slot-level schemes.}
  \label{fig_2}
  \vspace{-1.0em}
\end{figure}

Assuming neighboring STAs are legacy devices using BEB. 
In non-saturated traffic scenarios, the collision probability in a given slot correlates with the number of contending STAs. 
In saturated scenarios, it becomes a function of STAs' cumulative backoff countdown slots. For a legacy STA that has waited $b'$ generic slots, its conditional transmission probability becomes 
$\text{Pr}(\text{Transmit}| b') = \frac{1}{\text{CW}-b'}$
in the subsequent slot, where $\text{CW}$ denotes its current contention window size.
Each idle slot waited inherently elevates collision risks -- not only by decrementing neighbors' remaining backoff counters but also by allowing new contenders to emerge, unless a competitor transmits. 
Transmissions will either reduce contention or at least reset senders' conditional transmission probability. 

Motivated by this analysis, a binary backoff selection strategy can be proposed. STA selects a backoff counter of either 0 (immediate transmission, if defer period indicates low collision probability) or the current CW size (conservative deferral for the next opportunity).
We extend this method to probabilistic parameter transitions for multi-agent scenarios, minimizing collision rate in partially observable environments.

\subsection{Fairness Evaluation Metric for Backoff Selection}
The fairness of the existing enhanced DCA (EDCA) is determined by a set of configurable parameters (e.g., $\text{CW}_{\min}$, $\text{CW}_{\max}$, AIFS, TXOP limit). 
The presence of various services with different quality of service (QoS) requirements and traffic characteristics makes fairness quantification impractical, particularly in heterogeneous networks. 
General metrics, such as proportional fairness, may not be suitable for covering the diverse scenarios encountered in EDCA. 
Specifically, backoff selection mainly influences access opportunity fairness rather than channel occupancy, necessitating a dedicated evaluation criterion tailored to EDCA dynamics.

We propose a backoff ratio metric to assess fairness in backoff selection strategies. 
Let $B$ be a discrete random variable representing the cumulative backoff countdown value selected under a given strategy before transmission. By normalizing $B$ against the corresponding CW size, we define \textit{backoff ratio} as 
\begin{equation}
  \bar{B}\triangleq \frac{B}{\text{CW}}.
  \label{eq:backoff_ratio}
\end{equation}
A strategy is considered fair if the expectation of $\bar{B}$ satisfies 
\begin{equation}
  \mathbb{E}[\bar{B}] \ge 0.5.
  \label{eq:fairness}
\end{equation}
The proposed metric has several advantages: 
\emph{i)} The metric inherently reflects EDCA's BEB mechanism, where STAs randomly select backoff values from $[0,\text{CW}]$. The threshold 0.5 corresponds to the expected backoff ratio under ideal BEB operation.
\emph{ii)} The metric accommodates heterogeneous CW configurations across access categories (ACs), ensuring fairness evaluation respects prioritized QoS requirements.
\emph{iii)} Frequent collisions increase the average CW, penalizing poorly performing strategies by raising their effective $b$. Conversely, collision-free strategies achieve lower average backoff durations, rewarding higher throughput while maintaining fairness.

\subsection{Constrained MARL Problem}
In MARL scenarios, each agent makes decisions based on local observations, aiming to maximize its expected cumulative reward. The primary challenges in MARL arise from environmental non-stationarity and partial observability. 
To address these issues while minimizing additional communication overhead, the centralized training with decentralized execution (CTDE) paradigm has emerged as a widely adopted solution. 
Within this framework, we employ an actor-critic architecture, where the actor makes decentralized decisions using local observations, and the critic leverages global information for policy evaluation. 
From a fairness perspective, all STAs are considered equivalent and should employ identical access strategies. We therefore adopt shared parameters among agents, denoted by $\pi$ as the shared channel access policy.

The primary objective for channel access optimization is to minimize collision probability while ensuring fairness among STAs.
To formalize this problem, we extend the decentralized partially observable Markov decision process (Dec-POMDP) framework by incorporating fairness constraints. The resulting constrained Dec-POMDP is defined by the tuple $\langle \mathcal{M}, \mathcal{S}, \{\mathcal{A}^{(i)}\}, \{ \mathcal{O}^{(i)}\}, R, \{C^{(i)}\}, P, \gamma \rangle$.
Here, $\mathcal{M}$ denotes the set of agents with $|\mathcal{M}|=m$. 
$\mathcal{S}$ is the global state space.
$\mathcal{A}^{(i)}$ and $\mathcal{O}^{(i)}$ are the action space and observation space for agent-$i$, respectively. The joint action space is denoted as $\mathcal{A} = \prod\nolimits_{i} \mathcal{A}^{(i)}$.
$R: \mathcal{S}\times \mathcal{A} \mapsto \mathbb{R}$ specifies the reward function, and $C^{(i)}: \mathcal{S}\times \mathcal{A} \mapsto \mathbb{R}$ defines agent-specific constraint functions. 
$P:\mathcal{S}\times\mathcal{A} \times \mathcal{S} \mapsto [0,1]$ is the transition matrix, and $\gamma \in [0,1)$ is the discount factor. 

The discounted reward objective to be maximized for policy $\pi$ given the initial state distribution $\mu$ is
\begin{equation}
  J^{\pi}_{R} = \mathbb{E}^{\pi}_{s\sim \mu}\left[\sum_{t=0}^{\infty} \gamma^t r_t\right],
\end{equation}
where $r_t$ denotes the collision-avoidance reward at step $t$.
For a trajectory $\tau = (s_0, \boldsymbol{a}_0, s_1, \dots)$ with length $T$, the empirical backoff ratio constraint for agent-$i$ derived from (\ref{eq:fairness}) is
\begin{equation}
  \hat{\mathbb{E}}^{\pi} \left[\bar{B}^{(i)}\right] = \frac{1}{T_s^{(i)}} \sum_{k=0}^{T_s^{(i)}-1}\bar{b}_k^{(i)} \ge 0.5,
  \label{eq:constraint_mean}
\end{equation}
where $T_s^{(i)}$ counts agent-$i$'s transmissions in $\tau$, and $\bar{b}^{(i)}_k$ denotes the backoff ratio for $k$-th transmission. 
We design a stepwise penalty function 
\begin{equation}
  c^{(i)}_t \triangleq \left\{ 
    \begin{aligned}
      \frac{T}{T_s^{(i)}}\left(0.5-\bar{b}^{(i)}_t\right), &~\text{if agent-}i~\text{transmits at} ~t\\
      0, &~\text{otherwise.}
    \end{aligned}
  \right.
  \label{eq:constraint_violation}
\end{equation}
to allocate constraint credit temporally.
The expected mean valued constraint (\ref{eq:constraint_mean}) with respect to $\mu$ can be reformed as
\begin{equation}
  J_{C^{(i)}}^{\pi} = \lim_{T\rightarrow\infty}  \mathbb{E}^{\pi}_{s\sim \mu} \left[ \frac{1}{T} \sum_{t=0}^{T-1}c_t^{(i)}\right] \le 0.
  \label{eq:constraint_agent}
\end{equation} 
The constrained Dec-POMDP optimization problem becomes
\begin{equation}
  \max_{\pi} ~J_R^{\pi}, \quad \text{s.t.}~ J_{C^{(i)}}^{\pi} \le 0, ~\text{for}~i=1,\dots,m.
  \label{eq:problem}
\end{equation}

\section{Fairness-Constrained Channel Access\label{sec3}}
In this section, the proposed fairness-constrained MAPPO (FC-MAPPO) algorithm is introduced to solve the optimization problem in (\ref{eq:problem}), along with the designs of the Dec-POMDP components and network architecture.

\subsection{Design of Dec-POMDP Components}
{\bf Actions:} When the queue remains non-empty after a channel access deferral event, the agent selects a new backoff value $a^{(i)} \in \mathcal{A}^{(i)}$ based on local observations $o^{(i)}\in \mathcal{O}^{(i)}$. Given that the minimum CW size across all ACs is 3, we expand the binary backoff selection by defining the action space as $\{0, 1, 2, \text{CW}\}$.

{\bf Local Observations:} To estimate neighboring STAs' transmission probabilities at the next accessible slot, we adopt the delay to last successful transmission (D2LT) metric from \cite{guo2022multi}.
Additionally, we propose a novel metric called cumulative idle time since last observed ACK (CI2LA), calculated by subtracting the observed channel busy time from D2LT. This implementation requires protocol stack modifications for neighbor ACK monitoring.
For agent-$i$ observing $l$ active neighbors at step $t$, the local observation consists of two components:
\emph{i)} Neighbor observation: $o^{(-i)}_{t} \triangleq \{v^{(j)}_t, v'^{(j)}_t\}_{j=1}^l$, where $v^{(j)}_t$ and $v'^{(j)}_t$ are neighbor-$j$'s D2LT and CI2LA, respectively;
\emph{ii)} Self observation: $o^{(+i)}_{t} \triangleq [\text{CW}^{(i)}_t, d_t^{(i)},b'^{(i)}_t,f_t^{(i)},f'^{(i)}_t]$, where $d_t^{(i)}$ is the accumulated deferral times for the current access attempt, $b'^{(i)}_t$ is the cumulative backoff slots, $f_t^{(i)}$ and $f'^{(i)}_t$ are the delay and idle time since last transmission/reception failure, respectively.
The complete local observation for agent-$i$ is formally defined as $o^{(i)}_t \triangleq (o^{(+i)}_t, o^{(-i)}_t)$.

{\bf Global State:} We define the global state at step $t$ as $s_t \triangleq (O^{(+)}_t, O^{(-)}_t)$, where $O^{(+)}_t \triangleq \{o^{(+i)}_t\}_{i=1}^m$ is the set of the self observations of all agents and $O^{(-)}_t \triangleq \{v^{(j)}_t, v'^{(j)}_t\}_{j= 1}^{n-m}$ consists of the neighbor observations of the rest legacy STAs. 

{\bf Reward:} The reward function is designed as to mitigate collisions. We reward the successful transmissions and punish the cases without receiving an ACK. For MARL scenarios, to make agent learn how to cooperate, we encourage the agent with the largest ratio of cumulative backoff slot to its CW size to transmit in higher priority. 
Define the reward function as
\begin{equation}
  r_t \triangleq \left\{
  \begin{aligned}
    \frac{\bar{b}_t^{(i)}}{\bar{b}'_{t,\max}}, &~\text{if agent-}i ~\text{transmits successfully}, \\
    -2, &~\text{if agent-}i~ \text{transmits but no ACK},\\
    0, &~\text{otherwise},
  \end{aligned}
  \right.
  \label{eq:reward}
\end{equation}
where $\bar{b}'_{t,\max} = \max_{i=1,\dots,m} \frac{b'^{(i)}_{t}}{\text{CW}^{(i)}_{t}}$.

\subsection{Proposed FC-MAPPO Algorithm}
\begin{figure}[!t]
  \vspace{-1em}
  \centering
  \subfloat[Actor]{
    \includegraphics[width = 0.2\textwidth]{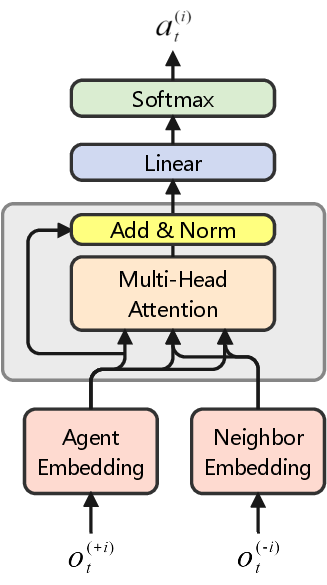}
  }
  \hspace{0.02\textwidth}
  \subfloat[Critic]{
    \includegraphics[width = 0.2\textwidth]{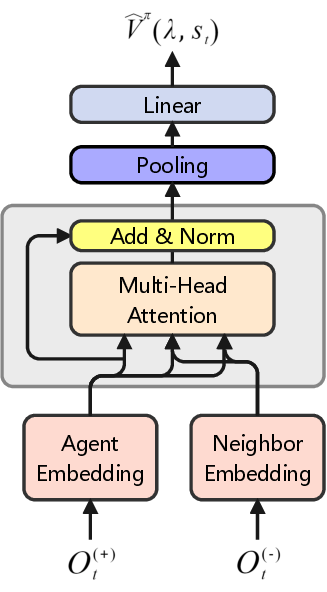}
  }
  \caption{MAPPO architecture: the actor selects action based on self and neighbor observations, while the critic collects local information to form global state for policy evaluation. }
  \label{fig_3}
\end{figure}

To handle the dynamic number of agents and implicit interactions among varying neighboring STAs sharing the channel, we employ a permutation-invariant \emph{transformer} architecture as shown in Fig. \ref{fig_3}. 
This design inherently captures hidden inter-STA dependencies through self-attention mechanisms while maintaining order invariance, enabling robust prediction of future collision probabilities.

Each agent's actor processes both active neighbors' MAC-layer statistics $o^{(-i)}$ and self-state information $o^{(+i)}$, concatenating them into a unified local observation $o^{(i)}$.
These features are processed through an attention layer where $o^{(+i)}$ serves as the query vector, and $o^{(i)}$ generates keys/values. The resulting environment-aware embedding informs distributed policy decisions without revealing neighbor identities.
During centralized training, the critic aggregates all agents' local observations at each trajectory step to construct global state.
The aggregated agent feautures $O^{(+)}$ function as queries combined with legacy neighbor features $O^{(-)}$ providing keys/values, followed by pooling layer that projects the resultant embeddings into fixed-dimensional representations.

While Lagrangian relaxation is widely adopted for constrained optimization, the mean-valued constraint in (\ref{eq:problem}) lacks the recursive property required for temporal difference (TD) learning in critic network.
The RCPO method \cite{tessler2018reward} addresses this by employing the discounted penalty under the assumption that local minima of modified constraints belong to the feasible set of the original constraint.
Meanwhile, to ensure constraint satisfaction, the Lagrangian multiplier $\lambda$ is still optimized using Monte Carlo sampling on the original constraint.

We adopt this framework and formulate the Lagrangian
\begin{equation}
  L(\lambda, \theta) = J^{\pi_{\theta}}_R- \lambda \sum_{i=1}^m \mathbb{E}^{\pi_{\theta}}_{s\sim\mu} \left[ \sum_{t=0}^{\infty} \gamma^t c^{(i)}_t  \right]
  \label{eq:lagrangian}
\end{equation}
yielding the unconstrained saddle-point problem 
\begin{equation}
    \min_{\lambda\ge 0} \max_{\theta} L(\lambda, \theta)
\end{equation},
where $\theta$ parameterizes the policy $\pi$, and the scaling factor $(1-\gamma)$ is absorbed into the Lagrange multiplier $\lambda$.
For homogeneous agents with identical constraints, a single $\lambda$ is used across all agents.
Define the penalized reward as 
\begin{equation}
  \tilde{r}(\lambda, s_t, a_t) \triangleq r_t - \lambda \sum_{i=1}^m c^{(i)}_t.
  \label{eq:penalized_reward}
\end{equation}
The MAPPO framework \cite{yu2021multi} is adapted by simply replacing original rewards with the penalized version.

\begin{algorithm}[!ht]
  \caption{Proposed FC-MAPPO Algorithm}
  \label{algorithm}
  \begin{algorithmic}[1]
    \STATE {\bf Initialize:} Lagrange multiplier $\lambda_0$, actor $\theta_0$, critic $v_0$, 
      learning rates $\eta_1(0)$, $\eta_2(0)$, $\eta_3(0)$, 
      simulation clock $t_{\text{sim}}=0$, total time $T_{\text{sim}}$, 
      step counter $t=0$.
    \FOR{$k=0,1,\dots$}
      \WHILE{$t_{\text{sim}} < T_{\text{sim}}$}
        \FOR{each agent-$i\in \mathcal{M}$}
          \STATE Collects local observation $o^{(i)}_{t}$.
          \IF{$t > 0$}
            \STATE Record last action result $r^{(i)}_{t-1}$.
            \STATE Calculate constraint violation $c^{(i)}_{t-1}$ as in (\ref{eq:constraint_violation}).
            \STATE Upload transition $(o^{(i)}_{t-1}, a^{(i)}_{t-1}, r^{(i)}_{t-1}, c^{(i)}_{t-1}, o^{(i)}_{t})$.
          \ENDIF
          \STATE Sampling action $a_{t}^{(i)} \sim \pi_{\theta_k}(\cdot|o^{(i)}_t)$.
        \ENDFOR
        \STATE $t \leftarrow t+1$
        \STATE $t_{\text{sim}} \leftarrow$ next\_access\_deferral\_event.timestamp.
      \ENDWHILE
      \STATE Construct global state $\{s_t\}_{t=0}^T$ and joint action $\boldsymbol{a}_t$.
      \STATE Calculate penalized reward $\{\tilde{r}_t\}_{t=0}^T$ as in (\ref{eq:penalized_reward}).
      \STATE Update actor-critic $\theta_{k+1}$, $v_{k+1}$ with $\{\tilde{r}_t\}_{t=0}^T$.
      \STATE {\bf Lagrange multiplier update:} $\lambda_{k+1}$ as in (\ref{eq:lambda_update}).
    \ENDFOR
  \end{algorithmic}
\end{algorithm}

The dual variable $\lambda$ is updated via gradient ascent on the original constraint violations in (\ref{eq:problem}) 
\begin{equation}
  \lambda_{k+1} \leftarrow \lambda_{k} + \eta_1(k)\sum_{i=1}^m \left(\frac{1}{T^{(i)}}\sum_{t=0}^{T^{(i)}-1}c_t^{(i)} \right),
  \label{eq:lambda_update}
\end{equation}
where $\eta_1(k)$, $\eta_2(k)$ and $\eta_3(k)$ denote the learning rates of $\lambda$, actor parameters $\theta$ and critic parameters $\phi$.
The procedure of the proposed algorithm is detailed in Algorithm \ref{algorithm}.

\section{Performance Evaluation\label{sec4}}

This section evaluates the performance of the proposed algorithm against conventional BEB strategy using an ns-3-based simulation platform adapted from \cite{pan2024macrevivoartificialintelligence}.
The simulation spans $T_{\text{sim}}=6$s, with the first second dedicated to BSS association establishment and the remaining five seconds for application traffic generation.
STAs generate saturated uplink traffic stream using 1500-byte packets under AC\_BE, configured with the ns-3 default parameters $\text{CW}_{\min}=15$ and $\text{CW}_{\max}=1023$.
To isolate collision-induced packet losses from channel impairments, STAs are randomly distributed within a 2-4 meter radius around the AP.
All nodes employ an 802.11ax protocol stack with fixed \textit{HE-MCS9} modulation for data frames.
The neural architectures (Fig. \ref{fig_3}) utilize single-head attention layer with uniform embedding dimensions of 64. 
Key training hyperparameters are detailed in Table \ref{table_simulation}.

\begin{table}[!ht]
  \caption{Main Parameters for simulation\label{table_simulation}}
  \centering
  \begin{tabular}{>{\Centering}p{2.5cm}|>{\Centering}p{0.6cm}|>{\Centering}p{2.8cm}|>{\Centering}p{1.1cm}}
    \toprule[1pt]
    \hline
    \textbf{Parameter} & \textbf{Value} & \textbf{Parameter} & \textbf{Value}\\
    \hline
        Discount factor $\gamma$ & 0.98 & Lagrange multiplier $\lambda$ & 0.01 \\
    \hline
        PPO epochs & 6 & Learning rate $\eta_1$ & $10^{-4}$ \\
    \hline
        PPO clip threshold & 0.2 & Learning rate (actor) $\eta_2$ & $3\times10^{-4}$ \\
    \hline
        GAE $\lambda_{\text{GAE}}$ & 0.95 & Learning rate (critic) $\eta_3$ & $10^{-2}$ \\
    \hline
    \bottomrule[1pt]
    \end{tabular}
    \vspace{-1.3em}
\end{table}

\subsection{Algorithm Training and Constraint Convergence}

\begin{figure*}[!t]
  \centering
	\subfloat[Reward per transmission]{
    \includegraphics[width=0.33\textwidth]{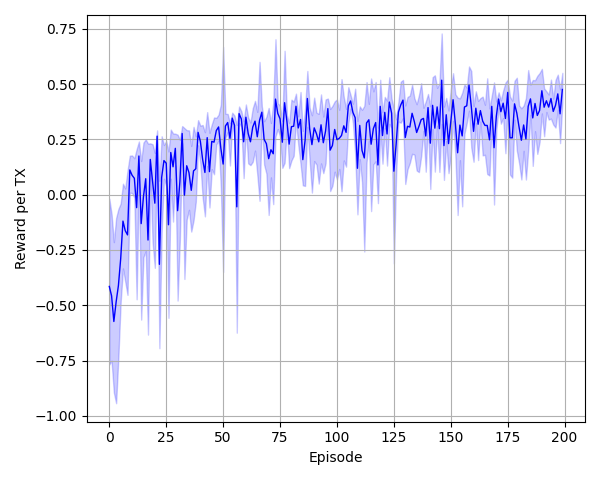}
    \label{fig_4a}
	}
  \hspace*{-0.8em}
  \subfloat[Backoff ratio]{
    \includegraphics[width=0.33\textwidth]{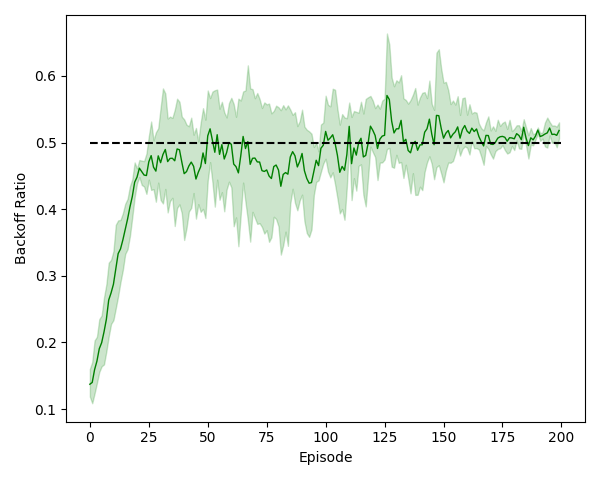}
    \label{fig_4b}
	}
  \hspace*{-0.8em}
  \subfloat[Lagrange multiplier]{
    \includegraphics[width=0.33\textwidth]{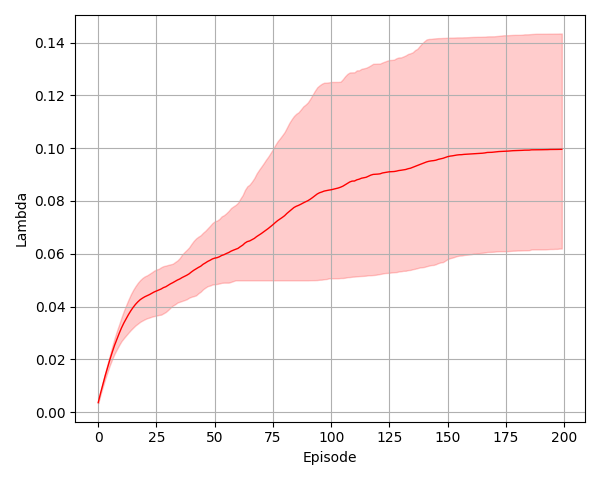}
    \label{fig_4c}
	}
  \caption{Convergence performance of the proposed FC-MAPPO algorithm with total number of STAs equal to 4.}
  \label{fig_4}
  \vspace{-1.3em}
\end{figure*}

Fig. \ref{fig_4} illustrates the learning dynamics of the proposed algorithm across 200 training episodes, averaged over 5 independent runs with $95\%$ boostrap confidence intervals. The training scenario fixes the total number of STAs to 4, with the count of RL-enhanced STAs randomly varying between 1 and 4 per episode to ensure robustness, while their positions are randomized. Fig. \ref{fig_4}\ref{sub@fig_4a} plots the per-transmission reward against episodes, demonstrating a rising trend with oscillations around 0.5 despite varying agent counts per episode. This indicates successful learning of collision avoidance strategies to improve transmission success rates. 
Fig. \ref{fig_4}\ref{sub@fig_4b} and \ref{fig_4}\ref{sub@fig_4c} depict the evolution of the average backoff ratio and the Lagrange multiplier $\lambda$, respectively. 
In early episodes, the agents prioritize constraint satisfaction to minimize penalties, leading to a rapid increase in the backoff ratio. 
As training progresses, the model gradually learns to reduce collisions while attempting to increase transmission frequency for higher rewards, even at the cost of occasional constraint violations.
This causes the backoff ratio to decline sharply. 
Through adaptive adjustments of $\lambda$, the penalty for constraint violations is intensified, ultimately driving the system to converge to a stable equilibrium, where constraints are satisfied while collision probabilities are minimized.

\subsection{Deployment Performance and Coexistence Analysis}
\begin{figure}[!t]
  \centering
	\includegraphics[width=0.48\textwidth]{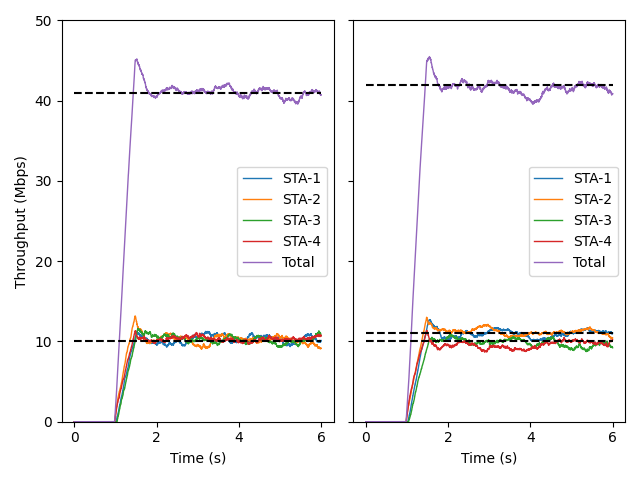}
    \vspace{-1.3em}
  \caption{Throughput comparison between homogeneous network (all 11ax) and heterogeneous deployment (half RL-enhanced).}
  \label{fig_5}
  \vspace{-1.3em}
\end{figure}

To evaluate the coexistence performance, Fig. \ref{fig_5} compares throughput between homogeneous network (4 11ax STAs) and heterogeneous scenario with STA 1\&2 employing policies trained in Fig. \ref{fig_4}. 
Results show that 11ax STAs maintain consistent throughput ($\sim$10Mbps), while RL-enhanced STAs achieve a $10\%$ throughput gain.
This collectively improves the overall system throughput by $5\%$.
The reduction in collision rates lower the average CW sizes of RL-enhanced STAs, securing more transmission opportunities. These additional chances would be wasted due to collisions in legacy-only scenarios, whereas RL-enhanced STAs efficiently exploit them. 

\begin{figure}[!t]
  \centering
  \includegraphics[width = 0.48\textwidth]{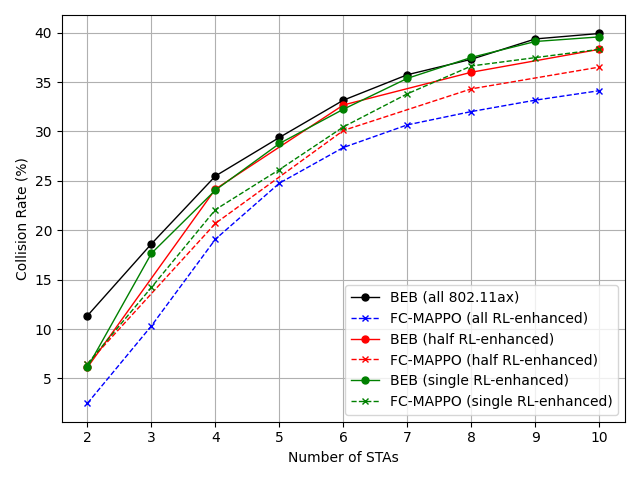}
  \vspace{-1.3em}
  \caption{Comparison of collision rates across different STA count and composition ratios.}
  \label{fig_6}
  \vspace{-1.3em}
\end{figure}

Fig. \ref{fig_6} further analyzes the average collision rates of the proposed FC-MAPPO algorithm and the baseline BEB strategy under varying STA population sizes. The BEB strategy suffers from rapidly escalating collision rates, reaching nearly $40\%$ with 10 STAs. 
In contrast, FC-MAPPO reduces collisions to $34\%$ in homogeneous scenarios. 
We also compare the average collision rates of legacy and RL-enhanced STAs under varying population ratios.
It reveals that higher proportion of RL-enhanced STAs lead to greater reductions in collision rates. 
Notably, even legacy STAs in heterogeneous networks experience a $2\%$ collision rate reduction, benefiting from the improved channel coordination of RL-enhanced STAs.


\section{Conclusions\label{sec5}}

This work demonstrates the viability of MARL in optimizing distributed channel access while harmonizing with legacy Wi-Fi systems. By integrating a dynamic backoff adaptation mechanism, EDCA-aligned fairness criteria, and decentralized constrained optimization, our framework achieves a 6\% reduction in collision probability compared to BEB while maintaining full backward compatibility. 
The proposed solution addresses critical limitations of conventional CSMA/CA in dense deployments. 
Future work will focus on generalizing the method to handle more complex and dynamic scenarios, specifically addressing variations in traffic and network size.


\bibliographystyle{IEEEtran}
\bibliography{ref}{}



\end{document}